%% file: geophysics_example.tex
\documentclass[noblind,paper,onecolumn]{geophysics}
\usepackage{makecell}
\usepackage{multirow}
\usepackage{amssymb}
\usepackage{amsmath}
\usepackage{booktabs}
\usepackage{hyperref}
\usepackage{subfig}

\begin{document}

\title{Hybrid Context-Fusion Attention (CFA) U-Net and Clustering for Robust Seismic Horizon Interpretation}

\renewcommand{\thefootnote}{\fnsymbol{footnote}} 



\address{
\footnotemark[1]Oxaala Tecnologias \\
Rua Dinah Silveira de Queirós, Nº 06, Quinta do Candeal, Horto Florestal, Salvador, Bahia, CEP 40.296.160 \\[0.4em]

\footnotemark[2]Geophysics Institute, Federal University of Bahia (UFBA) \\
R. Barão de Jeremoabo, Ondina, Salvador, Bahia, CEP 40170-290 \\[0.4em]

\footnotemark[3]Grupo de Estudo e Aplicação de Inteligência Artificial em Geofísica (GEAIG), Geophysics Institute, Federal University of Bahia (UFBA) \\[0.4em]

\footnotemark[4]Petrobras – Petróleo Brasileiro S.A. \\
Avenida República do Chile, 65, Centro, Rio de Janeiro, RJ, CEP 20031-912 \\[0.4em]

\footnotemark[5]Geological Survey of Brazil – Superintendência de Salvador \\
Av. Ulysses Guimarães, 2862, Sussuarana, Salvador, Bahia, CEP 41213-000
}

\author{
José Luis Lima de Jesus Silva\footnotemark[1]\footnotemark[3],
João Pedro Gomes\footnotemark[2],
Paulo Roberto de Melo Barros Junior\footnotemark[4],
Vitor Hugo Serravalle Reis Rodrigues\footnotemark[5],
and Alexsandro Guerra Cerqueira\footnotemark[1]\footnotemark[2]
}

\footer{J. Luis Silva, J. Gomes, P. Barros Junior, V. Rodrigues, and A. Cerqueira}
\lefthead{J. Luis Silva et al.}

\maketitle
\begin{center}
\textit{This is a preprint submitted to a journal. The final published version may differ.}
\end{center}

\makeatletter
\footnotetext{\seg@address} 
\makeatother
\begin{abstract}
Interpreting seismic horizons is a critical task for characterizing subsurface structures in hydrocarbon exploration. Recent advances in deep learning, particularly U-Net-based architectures, have significantly improved automated horizon tracking. However, challenges remain in accurately segmenting complex geological features and interpolating horizons from sparse annotations. To address these issues, a hybrid framework is presented that integrates advanced U-Net variants with spatial clustering to enhance horizon continuity and geometric fidelity. The core contribution is the Context Fusion Attention (CFA) U-Net, a novel architecture that fuses spatial and Sobel-derived geometric features within attention gates to improve both precision and surface completeness. The performance of five architectures, the U-Net (Standard and compressed), U-Net++, Attention U-Net, and CFA U-Net, was systematically evaluated across various data sparsity regimes (10-, 20-, and 40-line spacing). This approach outperformed existing baselines, achieving state-of-the-art results on the Mexilhão field (Santos Basin, Brazil) dataset with a validation IoU of 0.881 and MAE of 2.49~ms, and excellent surface coverage of 97.6\% on the F3 Block of the North Sea dataset under sparse conditions. The framework further refines merged horizon predictions (inline and cross-line) using Density-Based Spatial Clustering of Applications with Noise (DBSCAN) to produce geologically plausible surfaces. The results demonstrate the advantages of hybrid methodologies and attention-based architectures enhanced with geometric context, providing a robust and generalizable solution for seismic interpretation in structurally complex and data-scarce environments.
\end{abstract}
\newpage
\section{Introduction}

\input{001_Introduction}

\section{Dataset}

\input{002_dataset}

\section{Methods}

\input{003_Methods}

\section{Results}

\input{004_Results}

\newpage

\bibliographystyle{seg}  
\bibliography{example}

\end{document}

%% file: 001_Introduction.tex
Seismic interpretation is essential for characterizing subsurface geological formations and plays a key role in the oil and gas industry \citep{Mattos2021}. A critical aspect of this process is identifying key horizons within amplitude volumes. These horizons represent reliable, continuous reflection surfaces characterized by stable wavelet signatures in seismic surveys. High-fidelity mapping improves geological interpretation, allowing analysis of amplitude variations that can reveal important geological features. Therefore, accurate identification of these horizons is also essential to unraveling the temporal dynamics and formative processes that shape geological structures.

As seismic interpretation advances, the need for comprehensive three-dimensional (3D) data sets has become more apparent, especially in regions with complex geological structures. \cite{dorn1998modern} emphasized the limitations of traditional interpretations based on two-dimensional (2D) profiles. In large-scale surveys comprising numerous inlines and crosslines, manual horizon interpretation becomes labor-intensive and time-consuming. To mitigate this, advances in auto-tracking technology have emerged as viable alternatives \citep{Luo2023}. These tools are capable of tracing reflections iteratively using seed points and structural cues. However, their performance can be limited in complex settings, such as areas with faults or chaotic reflectivity patterns \citep{Wu2019}.

To overcome these limitations, a wide range of strategies have been proposed to enhance the automation of horizon tracking. \cite{Marfurt1999} introduced waveform similarity, which serves as a proxy for geological continuity, while \cite{Stark2003} proposed phase unwrapping to derive the Relative Geological Time (RGT), later extended with fault constraints by \cite{Wu2012}. Other methods utilize slope-based metrics and Dynamic Time Warping (DTW) to correlate seismic traces \citep{Hale2013, Wu2018}.

Initial attempts to integrate machine learning into horizon tracking explored multilayer perceptrons \citep{Harrigan1992, Kusuma1993}. However, the transformative impact came with modern deep learning techniques. Powered by advances in GPU computing and large-scale data sets, Convolutional Neural Networks (CNNs) have yielded significant improvements in waveform classification and pixel-wise segmentation \citep{LeCun2015, wu2018deep, peters2019multiresolution, Wu2019, yang2020seismic, calhes2021simplifying, ravasi2022joint}. Among these models, U-Net architectures \citep{ronneberger2015u}, with their encoder-decoder symmetry and skip connections, have proven particularly effective for segmentation of geological features \citep{Wu2019}.

Recent developments have expanded this paradigm to multi-scale 3D CNNs \citep{Tschannen2020} and advanced U-Net variants such as U-Net++ \citep{zhou2018unet++} and Attention U-Net \citep{alsalmi2024automated}. Despite progress, several critical challenges remain \citep{yu2021deep}, such as the robust interpolation of geological horizons from sparse annotations \citep{poulinakis2023machine}, and the precise segmentation of horizons in faulted and discontinuous environments, where signal ambiguity and structural complexity degrade performance.

Some of these challenges have been addressed by introducing and validating a comprehensive hybrid framework for seismic horizon interpretation, combining deep learning with spatial clustering. To enhance the performance in geologically complex regions,a cross-attention mechanism has been integrated based on the Attention U-Net \citep{oktay2018attention}, enabling the model to focus on salient seismic features. Additionally, a model-agnostic Density-based spatial clustering of applications with noise (DBSCAN) has been implemented as a post-processing step to filter out spurious and spatially incoherent predictions \citep{dhua2015segmentation}, thereby improving the geological plausibility of reconstructed surfaces.

The contributions are threefold. First, the Context-Fusion Attention (CFA) U-Net is proposed, a novel architecture that augments attention gates with spatial and Sobel-based heads to improve geometric precision and surface completeness. Second, a systematic comparative evaluation of five variants of the U-Net (Standard U-Net, compressed U-Net, U-Net++, Attention U-Net, and CFA U-Net) is conducted under varying annotation sparsity (10, 20, and 40-line spacing) in two geologically distinct data sets: the F3 Block in the North Sea Graben and the faulted Mexilhão Field in the Santos Basin ( Brazil). Third, the hybrid workflow, which combines attention mechanisms with DBSCAN-based clustering, yields more robust, accurate, and generalizable horizon interpretations than existing baselines.

The results confirm that the proposed Context-Fusion Attention (CFA) U-Net achieved the highest validation IoU ($0.881$) and lowest MAE ($2.49$~ms) on the faulted Mexilhão data set, outperforming all baselines. Furthermore, the model achieved the highest surface coverage ($97.6\%$) in sparse data scenarios, surpassing even the high-recall performance of U-Net++. These results highlight a measurable trade-off between precision and completeness, and demonstrate that attention-based architectures, when enhanced with context fusion mechanisms, offer a superior inductive bias for seismic interpretation under both sparse and geologically complex conditions.

%% file: 002_dataset.tex
\begin{figure}[!t]
    \centering
    \includegraphics[width=0.5\linewidth]{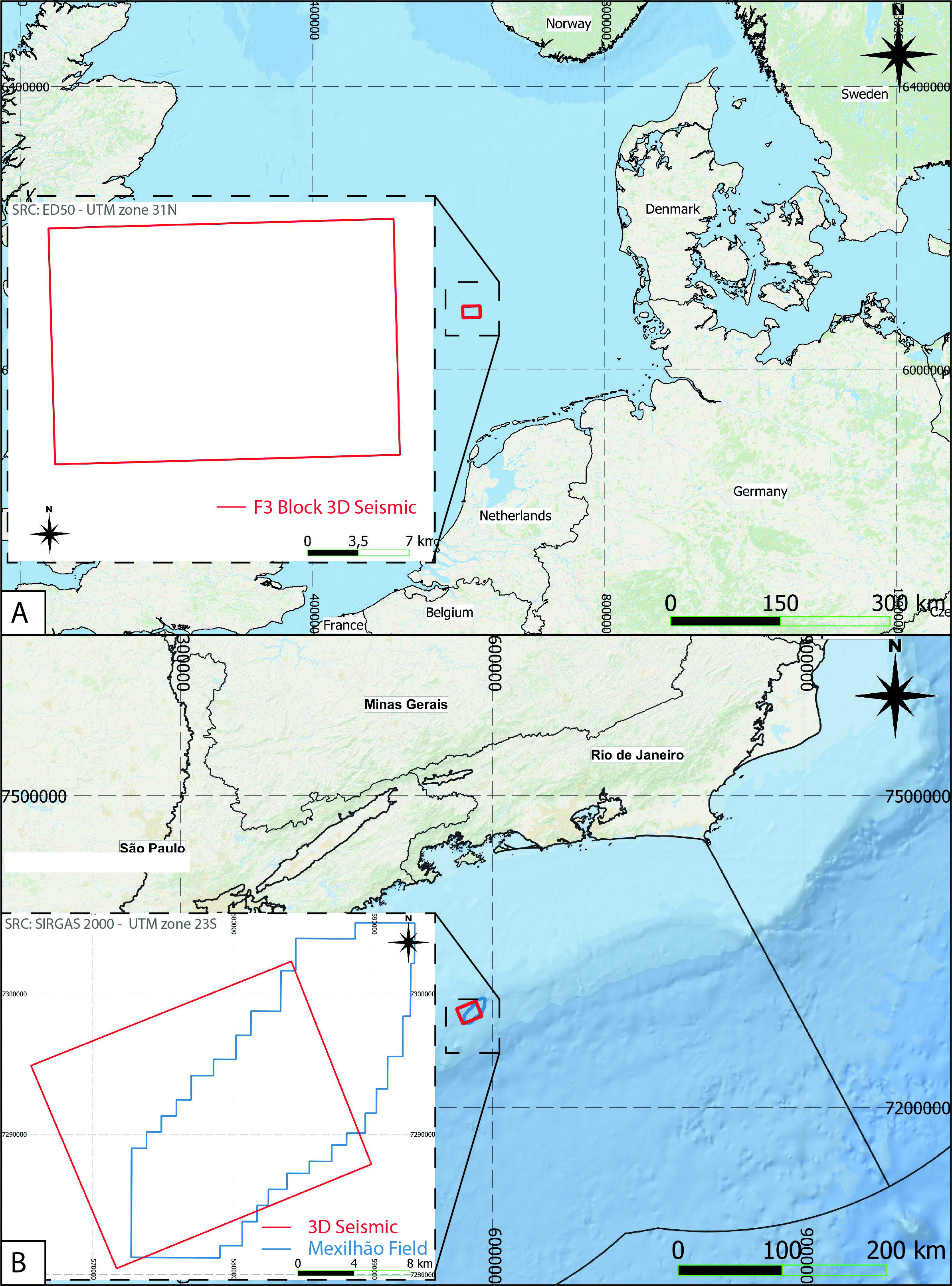}
    \caption{The two 3D seismic cubes utilized in the experiments are (a) the F3-Block in the Dutch North Sea and (b) the Mexilhão field in the Santos Basin, Brazil.}
    \label{fig:localization}
\end{figure}
To evaluate the performance of the methods on geologically diverse data, two distinct 3D seismic volumes were used, as shown in Figure~\ref{fig:localization}. The volume of seismic data F3 (survey from the Dutch sector of the North Sea) includes 651 inline and 951 crossline sections with a spatial sampling of 25 meters. The time dimension spans 1,848 ms, sampled at an interval of 4 ms. The FS8 seismic horizon is characterized by plane-parallel high-amplitude reflectors representing Cenozoic-era marine transgression deposits \citep{Schroot2005}. The reservoirs in this block exhibit clear direct hydrocarbon indicator (DHI) signatures \citep{Troccoli2022, Barbosa2024}.
\begin{figure}[!t]
    \centering
    \includegraphics[width=.7\linewidth]{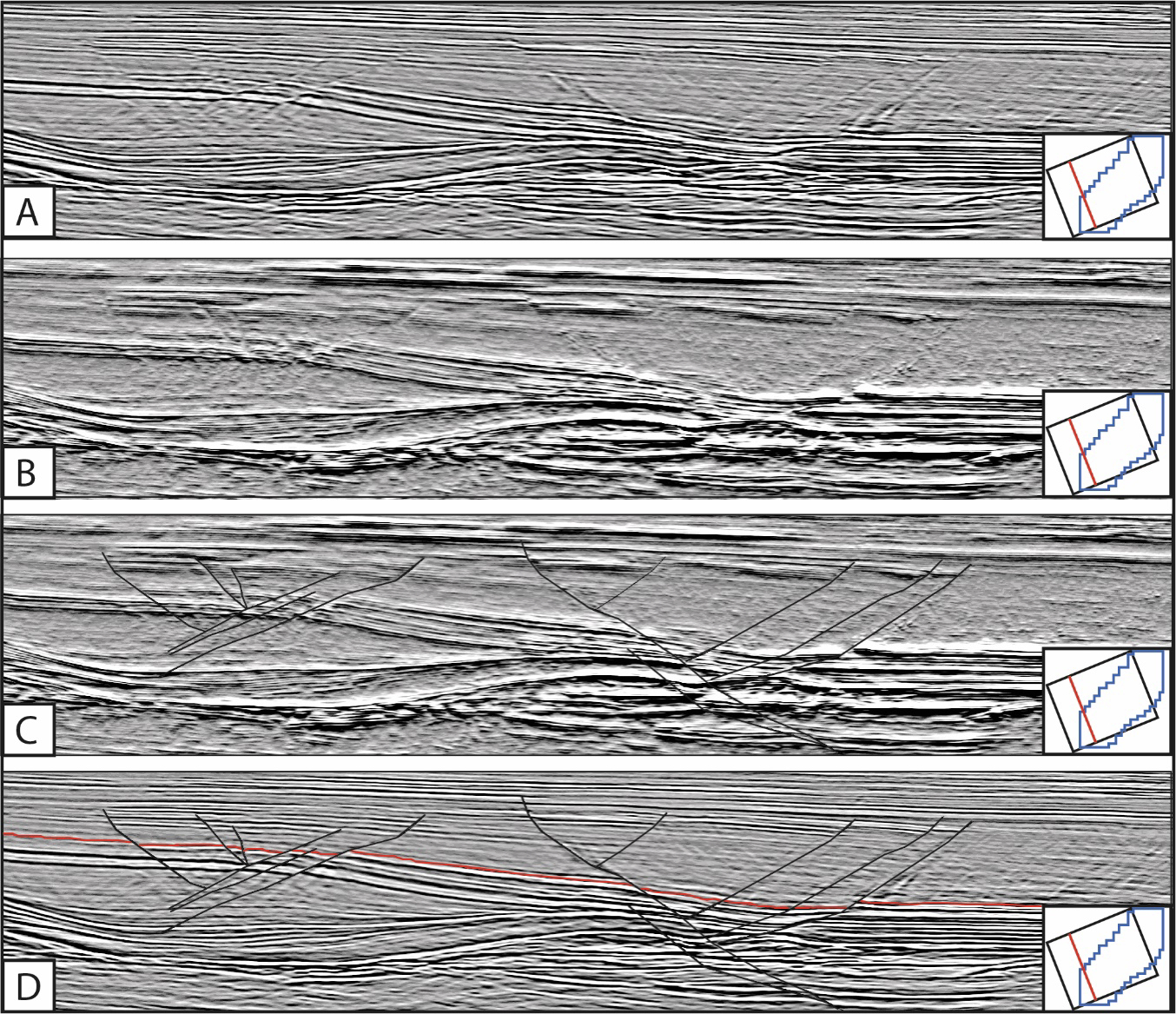}
    \caption{\textbf{Geological complexity of the Mexilhão field.} (a) A representative seismic amplitude slice. (b) The pseudo-relief attribute highlights structural features. (c) A manual fault interpretation overlaid on the pseudo-relief. (d) The seismic amplitude slice with both the fault interpretation and the target horizon highlighted.}
    \label{fig:mexilhao_data}
\end{figure}
The second data set (BS-400 block) covers the Mexilhão field in the Santos Basin (offshore of Brazil) which is renowned for its significant hydrocarbon reserves and natural gas. The Mexilhão field seismic volume presents a greater geological complexity, as illustrated in Figure~\ref{fig:mexilhao_data}. The data comprises 795 inlines and 624 crosslines, with a spatial resolution of 25 meters and a time range of 1600 ms, sampled at 4 ms. The geology is characterized by significant faulting resulting from halokinesis and extensional processes, which leads to reflectors with poor continuity in the rift section. Furthermore, the seismic data show good continuity in the reflectors of the drift portion. The largest reservoirs are turbiditic systems of the Ilha Bela Member, embedded within the shales of the Itajaí-Açu Formation. This complex structural and stratigraphic setting serves as an excellent test case for the robustness of the proposed segmentation framework.

To evaluate the models' ability to interpolate horizons from sparse interpretations, a systematic experiment was designed using both seismic cubes. For each of the six architectures, six distinct training scenarios were created by combining two orthogonal directions (inline and crossline) with three different sparsity levels, defined by the spacing between labeled lines (10, 20, or 40 lines). This resulted in a total of 36 unique trained models per seismic cube, allowing for a comprehensive assessment of each architecture's ability to generalize from sparse data. Table~\ref{tab:data sets} specifies the number of labeled lines used for training in each configuration.

\begin{table}[!h]
    \centering
    \begin{tabular}{llcc}
        \toprule
        \textbf{data set} & \textbf{Line Spacing} & \textbf{\# inlines} & \textbf{\# crosslines} \\
        \midrule
        \multirow{3}{*}[0.5ex]{F3} & 10 x 10 & 96 & 47 \\
        & 20 x 20 & 48 & 24 \\
        & 40 x 40 & 24 & 12 \\
        \cmidrule(lr){1-4}
        \multirow{3}{*}[0.5ex]{Mexilhão} & 10 x 10 & 63 & 225 \\
        & 20 x 20 & 32 & 113 \\
        & 40 x 40 & 16 & 57 \\
        \bottomrule
    \end{tabular}
    \caption{Number of labeled inlines and crosslines used for training in each sparse data configuration for the two seismic volumes.}
    \label{tab:data sets}
\end{table}

For each configuration, the data were partitioned using a systematic and non-random sampling strategy that directly simulates a real-world geophysical workflow. The sparsely sampled training set, for instance, consists of every 10th inline in the 3D volume. The remaining lines were reserved as the validation data set. From a machine learning perspective, this systematic partitioning serves as a form of spatial cross-validation \cite{bergmeir2012use}, which is crucial to evaluate models on spatially correlated data such as seismic volumes \citep{roberts2017cross}. This approach was deliberately chosen to avoid data leakage \citep{kaufman2012leakage} that could occur with a standard random shuffle. In geophysics, adjacent seismic lines are usually highly correlated. Therefore, the model trained on line $n$ and validated on the nearly identical line $n+1$ would yield artificially inflated performance metrics without truly testing the model's ability to generalize. In contrast, this approach rigorously tests the model's ability to interpolate geologically consistent horizons across large, unseen spatial gaps, which is the fundamental task required of an automated interpretation system.

%% file: 003_Methods.tex
\subsection{Architectural Framework}
\begin{figure}[!t]
    \centering
    \includegraphics[width=\linewidth]{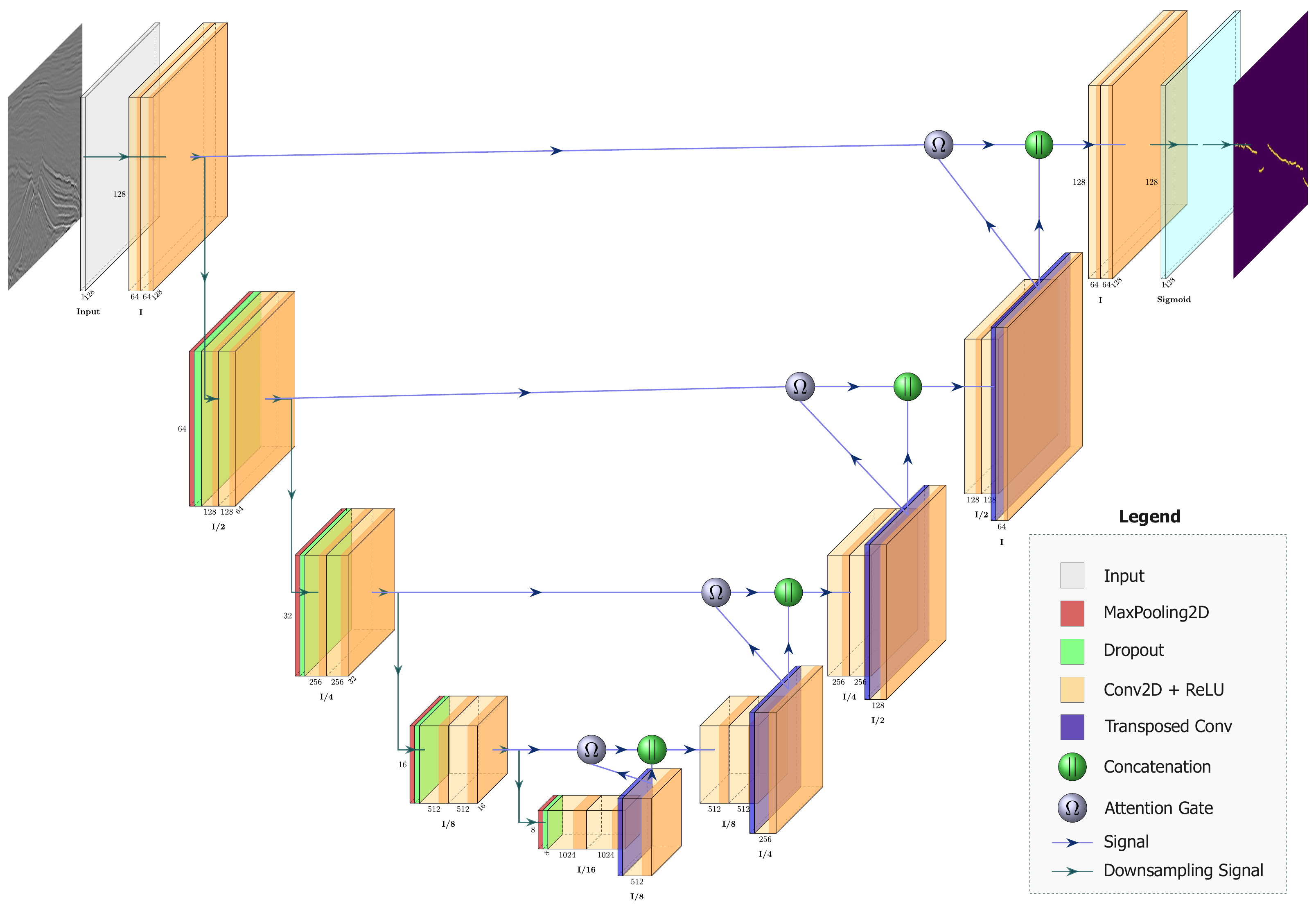}
    \caption{\textbf{Schematic of the Attention U-Net architecture.} The model processes a 2D seismic image patch (left) through a symmetric encoder-decoder network to produce a probability mask (right). The encoder (contracting path) uses convolutional (orange), dropout (green), and max-pooling (red) blocks to extract hierarchical features. The decoder (expansive path) uses upsampling (purple) and convolutional blocks to restore spatial resolution. The key components are the attention gates ($\Omega$) on the skip connections, which adaptively re-weight features before they are fused via concatenation ($||$).}
    \label{fig:model-flow}
\end{figure}
The segmentation of seismic horizons can be formulated as learning a mapping function $f: \mathbf{X} \to \mathcal{Y}$, and $\mathbf{X} \subset \mathbb{R}^{H \times W \times C}$ is the domain of 2D seismic image patches with height $H$, width $W$, and $C$ input channels, where $\mathcal{Y} \in \{0, 1\}^{H \times W}$ is the corresponding binary label space. To approximate this function, a family of deep convolutional neural networks derived from U-Net \citep{ronneberger2015u} is investigated. This foundational model is distinguished by its symmetric encoder-decoder structure with long-range skip connections that fuse deep contextual features with shallow fine-grained details, making it exceptionally effective for precise localization. A systematic analysis of several U-Net variants with significant differences in parameterization is presented, particularly their efficacy when trained on sparsely labeled data sets common in geophysical applications.

To investigate the impact of model capacity, a compressed U-Net variant \citep{wu2018deep} is included, which reduces the number of feature channels at each block. Furthermore, we explore two advanced architectures that modify the skip-connection mechanism. The U-Net++ architecture \citep{zhou2018unet++} introduces nested and dense skip pathways to bridge the semantic gap between the encoder and decoder. The Attention U-Net \citep{oktay2018attention} integrates learnable attention gates into the skip connections to adaptively re-weight encoder features. A general architecture diagram is illustrated in Figure~\ref{fig:model-flow} using the Attention U-Net. The network follows a symmetric encoder-decoder structure to process $128 \times 128$ single-channel seismic patches, producing a probability mask of the same dimension. The encoder (contracting path) progressively downsamples the input through four blocks to capture high-level features, while the decoder (expansive path) symmetrically upsamples the feature maps to reconstruct the full-resolution output. The key innovation of the Attention U-Net, the attention gate ($\Omega$), is integrated into each skip connection to filter features before they are concatenated ($||$) with the decoder path. The standard U-Net architecture is a simplification of this diagram where the attention gates are removed. 

Based on these architectures, a key limitation in standard attention mechanisms has been identified, leading to the proposal of the Context Fusion Attention U-Net ($CFA$ U-Net). The proposed model enhances the attention gate with a multi-head design that explicitly incorporates spatial and edge-aware inductive biases.

\subsection{Mathematical Formulation of Key Architectures}
As shown in the diagram Figure~\ref{fig:model-flow}, the Attention U-Net modifies the standard U-Net architecture consisting of a symmetric encoder-decoder structure linked by skip connections that are modulated by attention gates.

\subsubsection{The U-Net Encoder Pathway}
The encoder path progressively extracts hierarchical features and reduces spatial dimensions. Given an input image $\mathbf{z}^0 \in \mathbb{R}^{H \times W \times C}$, the encoder comprises $L$ levels. At each level $l \in \{1, \dots, L\}$, a convolutional block processes the input $\mathbf{z}^{l-1}$ to produce a feature map $\mathbf{x}_{\text{enc}}^l$, which is then passed to the corresponding decoder level via a skip connection, and then downsampled via max-pooling to produce the input for the next level \citep{krizhevsky2012imagenet}, $\mathbf{z}^l$, is found using the following expressions:
\begin{align}
    \mathbf{x}_{\text{enc}}^l &= \text{ConvBlock}^l_{\text{enc}}(\mathbf{z}^{l-1}) \label{eq:encoder_conv} \\
    \mathbf{z}^l &= \text{MaxPool}(\mathbf{x}_{\text{enc}}^l) \label{eq:encoder_pool}
\end{align}
where each $\text{ConvBlock}^l_{\text{enc}}$ consists of two sequential units, each comprising a $3 \times 3$ convolution, Batch Normalization \citep{ioffe2015batch}, and a Rectified Linear Unit (ReLU) activation function \citep{nair2010rectified}. The output of the final encoder level, $\mathbf{z}^L$, serves as the input to the bottleneck.

\subsubsection{Attention Gated Skip Connections}

To address the challenge of fusing semantically dissimilar features, Attention U-Nets \citep{oktay2018attention} employ attention gates (AGs) that adaptively re-weight the encoder feature maps to emphasize task-relevant features, $\mathbf{x}_{\text{enc}}^l$. The gating signal, $\mathbf{g}$, is the feature map from the next decoder level, $\mathbf{d}^{l+1}$, providing contextual information to guide the attention mechanism. The AG computes an attention coefficient map, $\mathbf{\alpha}^l \in [0, 1]^{H_l \times W_l}$:

\begin{equation}
    \mathbf{\alpha}^l = \sigma_2 \left( \psi \left( \sigma_1 \left( \theta_x(\mathbf{x}_{\text{enc}}^l) + \theta_g(\text{Up}(\mathbf{g})) \right) \right) \right)
    \label{eq:attention_gate}
\end{equation}
where $\theta_x(\cdot)$, $\theta_g(\cdot)$, and $\psi(\cdot)$ are linear projections implemented as $1 \times 1$ convolutions with their respective bias terms. The function $\sigma_1$ is a Rectified Linear Unit (ReLU) activation, and $\sigma_2$ is a sigmoid activation. The resulting attention-gated feature map, $\hat{\mathbf{x}}_{\text{enc}}^l$, is then computed via element-wise multiplication:
\begin{equation}
    \hat{\mathbf{x}}_{\text{enc}}^l = \mathbf{x}_{\text{enc}}^l \odot \mathbf{\alpha}^l
    \label{eq:gated_features}
\end{equation}
where $\hat{\mathbf{x}}_{\text{enc}}^l$ is the resulting \textbf{attention-gated feature map}, computed by performing an element-wise multiplication ($\odot$) between the original encoder feature map, $\mathbf{x}_{\text{enc}}^l$, and the attention coefficients, $\mathbf{\alpha}^l$. This operation effectively recalibrates the original features, emphasizing salient regions while suppressing irrelevant background information.

\subsubsection{Architectural differences in Skip Pathways}
In the standard U-Net \citep{ronneberger2015u}, the decoder receives features via a direct skip connection. At each level, the input to the convolutional block, $\text{ConvBlock}^l_{\text{dec}}$, is formed by concatenating the up-sampled feature map from the level below, $\text{Up}(\mathbf{d}^{l+1})$, directly with the original encoder feature map, $\mathbf{x}_{\text{enc}}^l$. This can be viewed as a simplification of the attention-based decoder, Equation~\ref{eq:decoder_block}, where the modulated feature map, $\hat{\mathbf{x}}_{\text{enc}}^l$, is replaced by the unmodified encoder features. This direct fusion combines deep, semantic information from the decoder path with shallow, high-resolution spatial details from the encoder path at each level.

The U-Net++ architecture \citep{zhou2018unet++}, conversely, redesigns the skip pathways to be nested and densely connected, aiming to bridge the semantic gap between the encoder and decoder feature maps more effectively. This is achieved by creating a series of intermediate convolutional blocks along the skip connections. Let $\mathbf{x}^{l,j}$ denote the output of the node at down-sampling level $l$ and convolutional layer $j$ of the dense skip path, where $l=0$ represents the top level and $j=0$ represents the output of the encoder. The output of any node $\mathbf{x}^{l,j}$ for $j > 0$ is computed by aggregating the outputs of all previous nodes at the same level, along with the up-sampled output from the corresponding node in the level below:
\begin{equation}
    \mathbf{x}^{l,j} = \text{ConvBlock}^{l,j}\left(\text{concat}\left[ \left( \mathbf{x}^{l,k} \right)_{k=0}^{j-1}, \text{Up}(\mathbf{x}^{l+1, j-1}) \right]\right) \quad \text{for } j > 0
    \label{eq:unetplusplus_skip}
\end{equation}
where $\text{ConvBlock}^{l,j}$ is the convolutional block at node $(l,j)$, consisting of the same operations (convolution, normalization, activation) as those in the encoder pathway, and $\text{concat}[\cdot]$ is the channel-wise concatenation operation. This formulation replaces the direct skip connection with a redesigned pathway where the decoder receives a densely aggregated set of feature maps. These maps are progressively enriched, facilitating a more gradual fusion of semantic and spatial information.

\subsubsection{Decoder Path}
The decoder path symmetrically restores the spatial resolution of the feature maps to that of the input image, while integrating the high-resolution, context-aware features from the attention-gated skip connections. The process begins with the feature map produced by the bottleneck, which we denote by $\mathbf{d}^{L+1}$. The decoder then proceeds through $L$ levels, from $l=L$ down to $1$. At each level $l$, the feature map from the previous (deeper) layer, $\mathbf{d}^{l+1}$, is first up-sampled by the operator $\text{Up}(\cdot)$. Upsampling is implemented as a learned $2 \times 2$ transposed convolution \citep{zeiler2014visualizing}, which allows the network to recover spatial information. The resulting tensor is then concatenated with the corresponding attention-gated feature map of the encoder, $\hat{\mathbf{x}}_{\text{enc}}^l$. This combined map is subsequently processed by a convolutional block of the decoder, $\text{ConvBlock}^l_{\text{dec}}$, to produce the output feature map $\mathbf{d}^l$:
\begin{equation}
    \mathbf{d}^l = \text{ConvBlock}^l_{\text{dec}} \left( \text{concat}\left[\text{Up}(\mathbf{d}^{l+1}), \hat{\mathbf{x}}_{\text{enc}}^l\right] \right)
    \label{eq:decoder_block}
\end{equation}
After the final decoder stage (at $l=1$), a $1 \times 1$ convolution is applied. This final layer acts as a pixel-wise classifier, mapping the multi-channel feature representation from the last decoder block to a single logit for each pixel. A sigmoid activation function is then applied to this output to produce the final probability map (horizon target) $\mathbf{\hat{y}} \in [0, 1]^{H \times W \times 1}$.
\subsubsection{Context-Fusion Attention U-Net (CFA U-Net)}
\begin{figure}[!t]
    \centering
    \includegraphics[width=\linewidth]{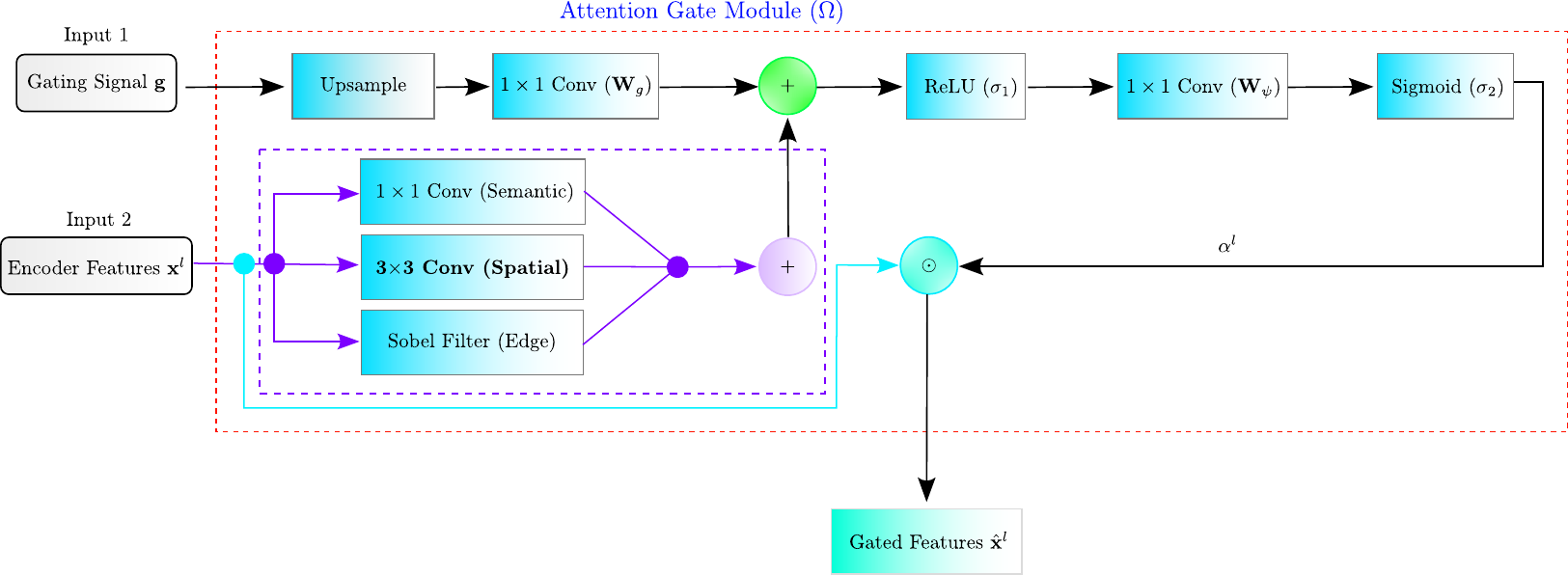}
    \caption{\textbf{Diagram of Context-Fusion Attention gate mechanism}}
    \label{fig:cfa-gate}
\end{figure}
Although the standard attention gate effectively reweighs features, its dependence on $1 \times 1$ convolutions makes it inherently spatially unaware. It determines the salience of features based on channel-wise information and cannot directly process local spatial patterns. This is a notable limitation for seismic interpretation, where key features are primarily defined by their local structure. To address this, the context-fusion attention gate, illustrated in Figure~\ref{fig:cfa-gate}, was designed to be sensitive to semantic content, learned spatial patterns, and explicit edge characteristics simultaneously.

The core innovation is the replacement of the standard encoder feature projection, $\theta_x(\mathbf{x}_{\text{enc}}^l)$ in Equation~\ref{eq:attention_gate}, with an enriched, multi-head representation, $\mathbf{h}_{\text{fused}}$. This is created by fusing the outputs of three parallel heads, where each acts as a specialized feature extractor:

First, a \textbf{Semantic Head} preserves the standard $1 \times 1$ convolution to identify the channel-wise importance of statistical properties, such as high seismic amplitude.
\begin{equation}
    \mathbf{h}_{\text{sem}} = \theta_{\text{sem}}(\mathbf{x}_{\text{enc}}^l)
    \label{eq:semantic_head}
\end{equation}

Second, a \textbf{Spatial Head} employs a $3 \times 3$ convolution to introduce a crucial \textbf{spatial inductive bias}, an architectural assumption that neighboring pixels are structurally related. This bias allows the model to recognize geometric forms, such as the continuity of a reflector, even where its statistical signature is weak.
\begin{equation}
    \mathbf{h}_{\text{spatial}} = \theta_{\text{spatial}}(\mathbf{x}_{\text{enc}}^l)
    \label{eq:spatial_head}
\end{equation}

Finally, an \textbf{Edge-Aware Head} provides a strong, \textbf{task-relevant prior} by incorporating a fixed-weight Sobel filter. By injecting explicit gradient information, this head frees the model from having to learn fundamental edge detection, allowing it to focus its capacity on learning the significance of those edges for the segmentation task.
\begin{equation}
    \mathbf{h}_{\text{edge}} = \theta_{\text{edge}}(\text{Sobel}(\mathbf{x}_{\text{enc}}^l))
    \label{eq:sobel_head}
\end{equation}

The outputs of these three heads from Equations \ref{eq:semantic_head}, \ref{eq:spatial_head}, and \ref{eq:sobel_head} are then fused via element-wise addition to create the following representation:
\begin{equation}
    \mathbf{h}_{\text{fused}} = \mathbf{h}_{\text{sem}} + \mathbf{h}_{\text{spatial}} + \mathbf{h}_{\text{edge}}
    \label{eq:fused_head}
\end{equation}
This fused tensor, $\theta_{x,fusion}=\mathbf{h}_{\text{fused}}$, directly replaces the $\theta_x(\mathbf{x}_{\text{enc}}^l)$ term in the attention mechanism. The attention coefficients for the CFA gate, $\mathbf{\alpha}_{\text{cfa}}^l$, are therefore computed by this direct substitution into Equation~\ref{eq:attention_gate}, such that:
\begin{equation}
    \mathbf{\alpha}_{\text{cfa}}^l = \sigma_2 \left( \psi \left( \sigma_1 \left( \mathbf{h}_{\text{fused}} + \theta_g(\text{Up}(\mathbf{g})) \right) \right) \right)
    \label{eq:cfa_gate}
\end{equation}

By endowing the attention gate with a multi-headed receptive field, the network learns not only \textit{what} features are relevant but also \textit{how they are structurally arranged}, significantly improving the segmentation of geologically complex regions. This design overcomes the spatial unawareness of standard attention gates. By creating three parallel processing streams, the fused knowledge enables the attention gate to make a more informed decision about not only which features are relevant but also how they are structurally arranged.

\subsubsection{Optimization and Evaluation}
To optimize the network parameters, a composite loss function $\mathcal{L}$ that combines the Dice Loss ($\mathcal{L}_{\text{Dice}}$) and Binary Cross-Entropy ($\mathcal{L}_{\text{BCE}}$) was employed. 
The BCE loss for a single pixel is given by:
\begin{equation}
    \mathcal{L}_{\text{BCE}}(y_i, \hat{y}_i) = - \left[ y_i \log(\hat{y}_i) + (1 - y_i) \log(1 - \hat{y}_i) \right]
\end{equation}
where $\hat{y}_i$ is the predicted probability for pixel $i$ and $y_i$ is the corresponding ground truth label ($0$ or $1$).
The Dice Loss is particularly effective for handling class imbalance in segmentation tasks and is defined as:
\begin{equation}
    \mathcal{L}_{\text{Dice}} = 1 - \frac{2 \sum_{i=1}^{N} y_i \hat{y}_i + \epsilon}{\sum_{i=1}^{N} y_i^2 + \sum_{i=1}^{N} \hat{y}_i^2 + \epsilon}
    \label{eq:dice_loss}
\end{equation}
where $N$ is the total number of pixels, and $\epsilon$ is a small constant for numerical stability. The final loss function is a weighted sum, $\mathcal{L} = \alpha \mathcal{L}_{\text{BCE}} + \beta \mathcal{L}_{\text{Dice}}$, with parameters $\alpha=0.5$ and $\beta=0.5$. Model performance was evaluated using the Intersection over Union (IoU) metric, or Jaccard index, defined as:

\begin{equation}
    \text{IoU} = \frac{|\mathbf{Y} \cap \mathbf{\hat{Y}}|}{|\mathbf{Y} \cup \mathbf{\hat{Y}}|} = \frac{\text{TP}}{\text{TP} + \text{FP} + \text{FN}}
    \label{eq:iou_metric}
\end{equation}

Where $\mathbf{Y}$ and $\mathbf{\hat{Y}}$ are the ground truth and predicted segmentation masks, and TP, FP, and FN are the true positives, false positives, and false negatives, respectively.

\subsection{DBSCAN-based Filtering of Horizon Predictions}

To refine the raw output from the neural networks and remove spurious, high-probability voxels that are geologically inconsistent with a continuous horizon, we employ the Density-Based Spatial Clustering of Applications with Noise (DBSCAN) algorithm as a postprocessing step \citep{ester1996density,deng2020dbscan}. DBSCAN is exceptionally well-suited for this task as it can identify clusters of arbitrary shape and effectively isolate noise points in low-density regions without requiring a predefined number of clusters \citep{schubert2017dbscan}.

The behavior of the DBSCAN algorithm is governed by two key parameters: the maximum neighborhood distance, $\epsilon$, and the minimum number of points required to form a dense region, \textit{MinPts}. Based on these parameters, each point in the data set is categorized according to its density. A point is classified as a \textbf{core point} if at least \textit{MinPts} neighbors are found within its $\epsilon$ radius. A \textbf{border point}, in contrast, is not a core point itself but falls within the $\epsilon$-neighborhood of one. Any point that satisfies neither of these conditions is subsequently labeled as a \textbf{noise point}. The algorithm forms final clusters by identifying connected components of core points and their associated border points, while all points designated as noise are discarded from the final result.

The input to the DBSCAN algorithm is a 3D point cloud generated from the probability volume, $\mathbf{\hat{y}}$, predicted by each architecture. To create the point cloud, a probability threshold of $\tau=1 \times 10^{-5}$ was first applied to the volume. The spatial coordinates $(i, j, k)$ of all suprathreshold voxels, where $\hat{y}_{ijk} > \tau$, were then extracted to form the point cloud for clustering.

The selection of the DBSCAN hyperparameters, `epsilon` ($\epsilon$) and `MinPts`, was determined empirically to suit the density of the predicted seismic data. The `MinPts` value was set to 25 to ensure that only geologically significant point groupings were considered as clusters. The `epsilon` parameter was calculated based on a vertical exaggeration factor ($z_{\text{factor}}=3$), which results in an effective neighborhood distance of $\epsilon = 6.0$ for clustering. After applying DBSCAN with these parameters, the algorithm identifies all clusters in the data. The final step of the filtering process involves retaining only the single most significant point cluster, which is presumed to be the actual horizon, and discarding all smaller clusters and noise points.

\subsection{Hybrid Segmentation Workflow}
\begin{figure}[!t]
    \centering
    \includegraphics[width=\linewidth]{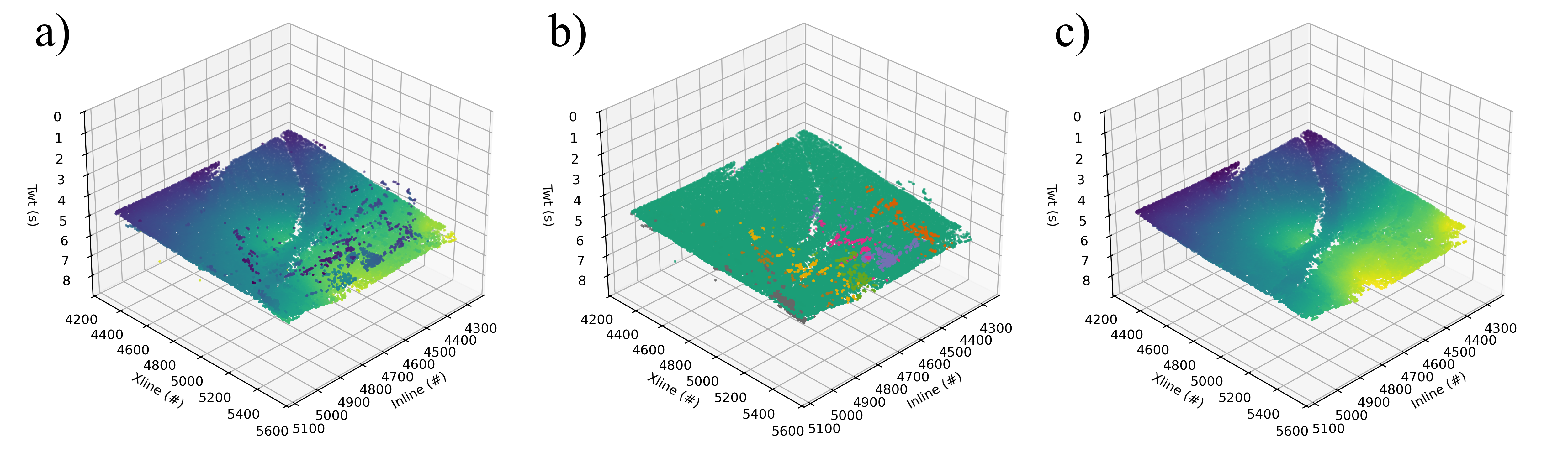}
    \caption{\textbf{The DBSCAN filtering workflow.} (a) The raw 3D probability volume is predicted by the neural network, with points colored by their two-way travel time (TWT). (b) DBSCAN groups the data into distinct clusters, identified by different categorical colors. (c) The final result after filtering for the largest point cluster, with the TWT-based color map reapplied to the cleaned horizon.}
    \label{fig:dbscan}
\end{figure}

\begin{figure}[!t]
    \centering
    \includegraphics[width=\linewidth]{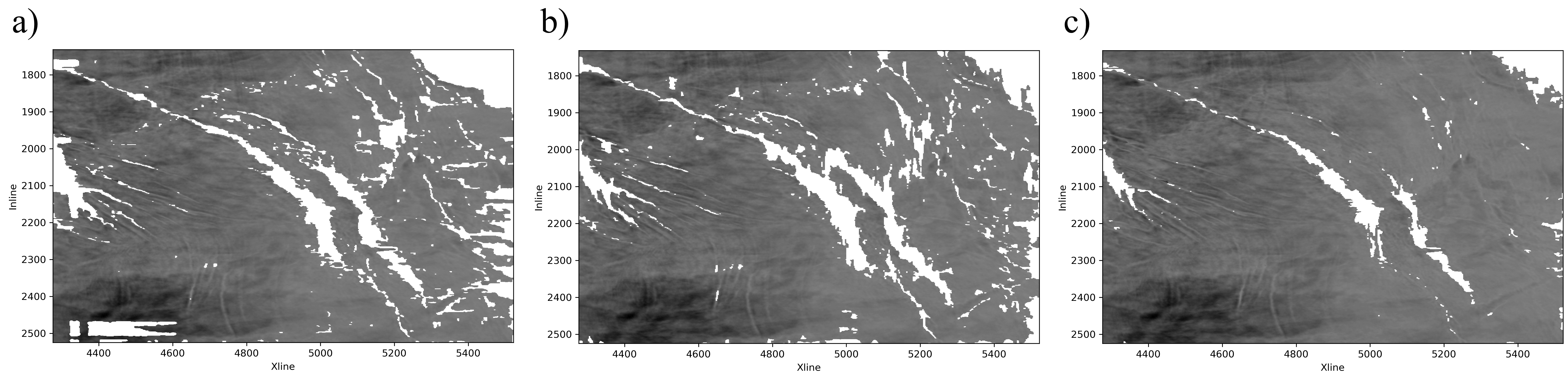}
    \caption{\textbf{Orthogonal prediction fusion.} This top-down view shows that predictions from models trained on \textbf{(a)} inline and \textbf{(b)} crossline directions are complementary. \textbf{(c)} represents the final surface with significantly improved completeness.}
    \label{fig:merging}
\end{figure}

\begin{figure}[!t]
    \centering
    \includegraphics[width=\linewidth]{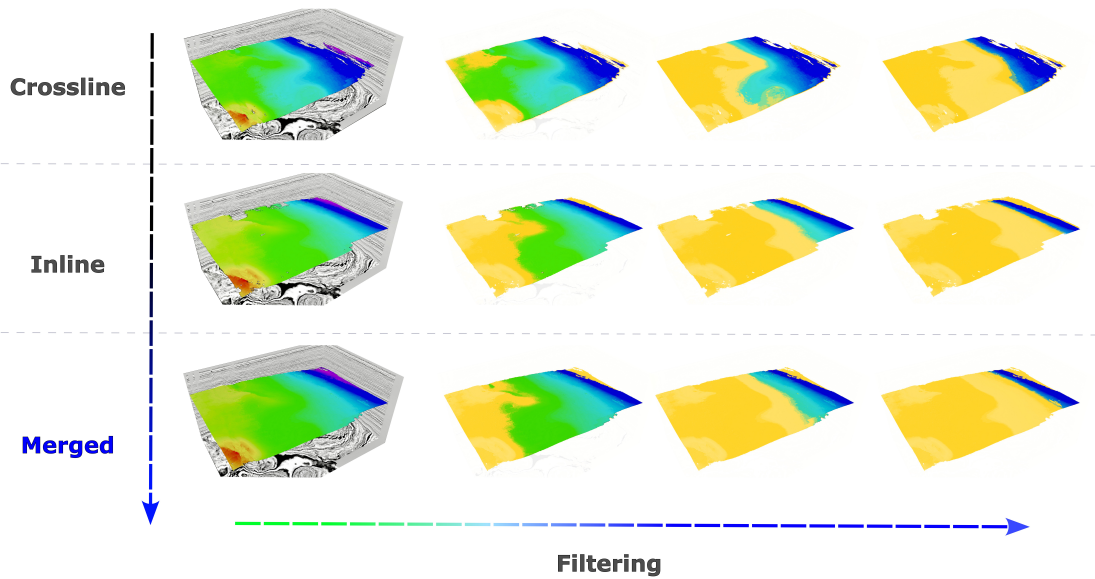}
    \caption{\textbf{Qualitative analysis of the orthogonal fusion workflow.} The top two rows show complementary predictions from models trained on crossline and inline directions, respectively. The bottom row shows the merged union, which produces a more complete horizon. The horizontal axis represents a progressive filtering of the surface by depth to visualize continuity.}
    \label{fig:3d_pred_example}
\end{figure}

The methodology for seismic horizon segmentation follows a multistage process that integrates deep learning for initial prediction with density-based clustering for postprocessing and refinement, as illustrated in Figures~\ref{fig:dbscan}, \ref{fig:merging}, and \ref{fig:3d_pred_example}. The DBSCAN filtering workflow is composed of three steps, as shown in Figure~\ref{fig:dbscan}. First, \textbf{(a)} the raw 3D probability volume is displayed as a point cloud where the color represents the two-way travel time (TWT), ranging from shallow (yellow) to deep (purple). However, the prediction contains significant and spatially isolated noise. Therefore, \textbf{(b)} the points are clustered by distinct categorical colors, where the main horizon becomes a large cluster (green) and the noise is separated into smaller clusters of different colors. Finally, \textbf{(c)} only the most significant cluster is retained, producing a clean surface where the TWT color map is reapplied to show the final geologically coherent structure. This postprocessing is part of the final stage of the hybrid workflow, which is composed of three primary stages: experimental design for training on sparse data, model implementation and training, and the inference and filtering pipeline.

A systematic evaluation of the models was performed under varying data sparsity conditions. For each of the six architectures, six different training scenarios were created by combining two orthogonal interpretation directions (inline and crossline) with three label spacings (10th, 20th, or 40th labeled line). This experimental design resulted in a total of 36 unique trained models, allowing a comprehensive assessment of the ability of each architecture to interpolate between sparsely labeled seismic lines. A systematic, nonrandom split was employed in which using every 10th, 20th, or 40th line resulted in training sets that comprise approximately 10\%, 5\%, and 2.5\% of the total available data, respectively. The remaining 90\%, 95\%, and 97.5\% lines, respectively, were retained for validation and testing.

All models were implemented in Tensorflow/Keras and trained on an NVIDIA 4080 GPU using the Adam optimizer \citep{kingma2014adam}. While a common procedure was followed, key hyperparameters such as learning rate and batch size were tuned for each architecture based on performance on a dedicated validation set, with the following learning rate (LR) and batch size (BS) pairs: 
\textbf{U-Net} (LR: $1 \times 10^{-4}$, BS: 1), \textbf{U-Net++} (LR: $5 \times 10^{-3}$, BS: 1), \textbf{Compressed U-Net} (LR: $5 \times 10^{-4}$, BS: 5), and \textbf{Attention U-Net} (LR: $5 \times 10^{-4}$, BS: 1). For the compressed U-Net, L2 regularization with a factor of $1 \times 10^{-4}$ was applied to convolutional kernels to mitigate overfitting. Models were trained for up to 500 epochs using the DICE loss function, and an early stopping callback that monitored validation loss and ceased training if no improvement was observed for 30 consecutive epochs. The architectural framework was designed to be flexible for input tensor dimensions, allowing models to be adapted to different data set resolutions through a programmatic shape optimization utility.

After training, each model generated a 3D probability volume, $\mathbf{\hat{y}}$, which was refined using a two-stage postprocessing pipeline. The DBSCAN was applied to filter the raw probability volume and retain only the most significant and densely connected cluster of points. This step effectively removes spurious predictions that are spatially isolated from the main horizon structure, as illustrated in Figure~\ref{fig:dbscan}.
To maximize spatial coverage and mitigate directional biases, the final step fused the filtered predictions from models trained on orthogonal inline and crossline data, which is an effective technique in volumetric image analysis \citep{stollenga2015parallel}. The fusion is performed by taking the set union of the two filtered 3D point clouds, as illustrated in Figure~\ref{fig:merging}:
\begin{equation}
    \mathbf{P}_{\text{final}} = \mathbf{P}_{\text{inline}} \cup \mathbf{P}_{\text{xline}}
    \label{eq:union}
\end{equation}
where $\mathbf{P}$ represents a filtered point cloud for a given model and spacing. This process combines the strengths of both orthogonal predictions to produce a more complete and geologically coherent horizon surface.

Figure~\ref{fig:3d_pred_example} shows the predicted horizon surfaces colored by the two-way travel time (TWT) \citep{sheriff1995exploration} (red is shallow, purple is deep). Using the U-Net++ architecture as an example, the figure highlights the layout that highlights the complementary nature of the orthogonal predictions. The rows compare the separate inline and crossline predictions against the final merged surface. Furthermore, the columns show a progressive filtering of the surface by depth, a technique that simplifies the identification of gaps and discontinuities.

The models' complementary performance is a direct result of interpreting three-dimensional, often anisotropic, geology using 2D seismic slices. For instance, the inline-trained model accurately captures the right side of the volume where the crossline model is weaker, while the crossline model resolves gaps in the upper surface that the inline model misses. By merging the two predictions via a set union given by Equation~\ref{eq:union}, the distinct strengths of each view-dependent interpretation are combined. Because a geological feature can have a clear signature in one view but appear as an ambiguous pattern in another, the merged horizon is more complete and geologically robust than an interpretation from a single orientation.

%% file: 004_Results.tex
\subsection{F3-Block Segmentation Results}
\begin{table}[!t]
    \centering
    \begin{tabular}{|l|c|cc|cc|cc|}
        \toprule
        \hline
        \textbf{Architecture} & \textbf{Spacing} & \multicolumn{2}{c|}{\textbf{Mean Acc.}} & \multicolumn{2}{c|}{\textbf{Mean IoU}} & \multicolumn{2}{c|}{\textbf{Mean Dice Loss}} \\
        & & Train & Valid & Train & Valid & Train & Valid \\
        \hline
        \midrule
        \multirow{3}{*}{U-Net (Comp.)} & 10 & 0.9979 & 0.9942 & 0.9635 & 0.9039 & 0.3280 & 0.1727 \\
        & 20 & 0.9971 & 0.9936 & 0.9496 & 0.8947 & 0.1343 & 0.1890 \\
        & 40 & 0.9961 & 0.9901 & 0.9312 & 0.8461 & 0.2472 & 0.2364 \\
        \hline
        \multirow{3}{*}{U-Net} & 10 & 0.9992 & 0.9970 & 0.9843 & 0.9458 & 0.0278 & 0.0741 \\
        & 20 & 0.9989 & 0.9970 & 0.9803 & 0.9464 & 0.0332 & 0.0740 \\
        & 40 & 0.9981 & 0.9958 & 0.9653 & 0.9268 & 0.0614 & 0.0792 \\
        \hline
        \multirow{3}{*}{U-Net++} & 10 & 0.9995 & 0.9963 & 0.9901 & 0.9331 & 0.0218 & 0.0876 \\
        & 20 & 0.9993 & 0.9967 & 0.9872 & 0.9403 & 0.0289 & 0.0816 \\
        & 40 & 0.9990 & 0.9953 & 0.9820 & 0.9181 & 0.0603 & 0.0941 \\
        \hline
        \multirow{3}{*}{Attn. U-Net} & 10 & 0.9996 & 0.9954 & 0.9931 & 0.9177 & 0.0197 & 0.1006 \\
        & 20 & 0.9993 & 0.9946 & 0.9865 & 0.9032 & 0.0288 & 0.1337 \\
        & 40 & 0.9990 & 0.9916 & 0.9812 & 0.8510 & 0.0712 & 0.1990 \\
        \hline
        \multirow{3}{*}{$CFA^{S}$ U-Net} & 10 & 0.9995 & 0.9951 & 0.9897 & 0.9127 & 0.0224 & 0.1174 \\
        & 20 & 0.9992 & 0.9947 & 0.9859 & 0.9056 & 0.0279 & 0.1214 \\
        & 40 & 0.9989 & 0.9917 & 0.9795 & 0.8571 & 0.1111 & 0.2360 \\
        \hline
        \multirow{3}{*}{$CFA$ U-Net} & 10 & 0.9989 & 0.9966 & 0.9796 & 0.9381 & 0.0332 & 0.0843 \\
        & 20 & 0.9980 & 0.9942 & 0.9631 & 0.8974 & 0.0521 & 0.1401 \\
        & 40 & 0.9983 & 0.9952 & 0.9677 & 0.9165 & 0.0650 & 0.1016 \\
        \hline
        \bottomrule
    \end{tabular}
    \caption{\textbf{Segmentation metrics on the F3 data set.} "Valid" refers to the validation set. Values represent the mean of models trained on inline and crossline data. IoU (higher is better) and Dice Loss (lower is better) are the most informative metrics for this unbalanced segmentation task.}
    \label{tab:metrics-f3}
\end{table}
The initial evaluation was performed on the F3 seismic volume, using the well-defined FS8 horizon as the segmentation target. Table~\ref{tab:metrics-f3} presents the mean training and validation metrics for each architecture across the three data sparsity configurations. All models achieved high accuracy scores (often $>0.99$), however, this metric can be misleading in segmentation tasks with severe class imbalance, where the background (non-horizon) class constitutes the vast majority of pixels \citep{jadon2020survey}. Therefore, the Intersection over Union (IoU) and Dice Loss provide more robust measures of performance. The results indicate that the standard U-Net architecture has strong and consistent performance, achieving both the highest validation IoU and the lowest validation Dice Loss across all data spacings, peaking with an IoU of $0.946$ and a loss of $0.074$ at the 10- and 20-line spacings. The proposed \textbf{$CFA$ U-Net} also demonstrated highly competitive performance, ranking second in validation IoU on the 10-line spacing data set with a validation IoU of $0.938$.

Following the pixel-level evaluation, the practical utility of each model is framed by the classic precision-recall trade-off \citep{davis2006relationship}, with results summarized in Table~\ref{tab:results-f3}. Precision is defined as geometric accuracy, measured by the Mean Absolute Error (MAE) and Mean Squared Error (MSE). Lower values for these metrics indicate a more precise reconstruction, with predicted points being more faithful to the actual reflector depth. Recall is defined as spatial completeness, measured by the percentage of surface coverage area. Higher values indicate a more complete reconstruction of the entire surface. This framework allows for a precise analysis of how each architecture manages the inherent tension between these competing objectives.
\begin{table}[h!]
    \centering
    \begin{tabular}{lccccc}
        \toprule
        \textbf{Architecture} & \textbf{Spacing} & \textbf{MAE} & \textbf{MSE} & \textbf{Area (\%)} \\
        \midrule
        \multirow{3}{*}{U-Net (Comp.)} & 10 & 4.21 & 22.69 & 96.34 \\
        & 20 & 4.20 & 22.69 & 95.27 \\
        & 40 & 4.35 & 25.86 & 91.74 \\
        \hline
        \multirow{3}{*}{U-Net} & 10 & 4.23 & 23.26 & 98.27 \\
        & 20 & 4.25 & 24.00 & 98.20 \\
        & 40 & 4.38 & 28.71 & 97.40 \\
        \hline
        \multirow{3}{*}{U-Net++} & 10 & 4.19 & 22.38 & 97.92 \\
        & 20 & 4.22 & 23.06 & 98.11 \\
        & 40 & 4.27 & 24.06 & 97.32 \\
        \hline
        \multirow{3}{*}{Attn. U-Net} & 10 & 4.14 & 21.39 & 93.26 \\
        & 20 & 4.18 & 22.31 & 91.63 \\
        & 40 & 4.21 & 23.04 & 86.77 \\
        \hline
        \multirow{3}{*}{\textbf{$CFA^{S}$ U-Net}} & 10 & 4.77 & 45.97 & 96.83 \\
        & 20 & 5.05 & 67.30 & 96.17 \\
        & 40 & 5.18 & 54.27 & 93.16 \\
        \hline
        \multirow{3}{*}{\textbf{$CFA$ U-Net}} & 10 & 4.44 & 33.99 & 98.19 \\
        & 20 & 4.65 & 43.11 & 95.92 \\
        & 40 & 4.83 & 47.83 & 97.61 \\
        \bottomrule
    \end{tabular}
    \caption{This table shows the geometric errors (MAE, MSE) and surface coverage area of the merged and filtered results for each model in the F3 data set. The proposed models are $CFA^{S}$ U-Net (semantic and spatial heads) and the $CFA$ U-Net (semantic, spatial, and Sobel heads).}
    \label{tab:results-f3}
\end{table}

\begin{figure}[h!]
    \centering
    \includegraphics[width=\linewidth]{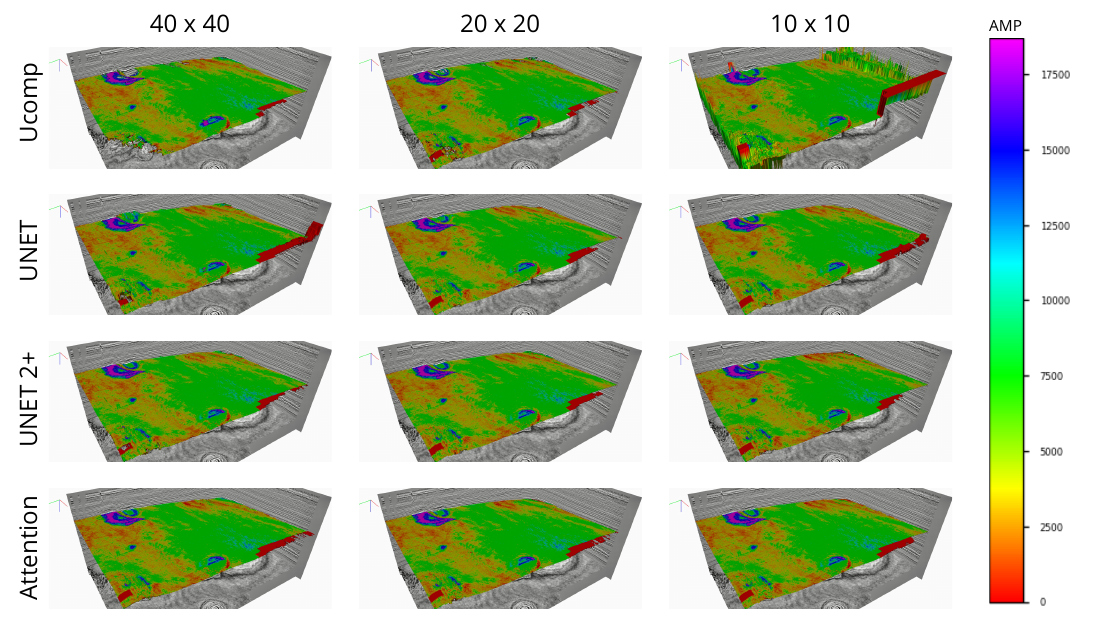}
    \caption{\textbf{Reconstructed 3D horizon surfaces after DBSCAN filtering for all architectures and data sparsity levels.} Each panel displays the final surface generated by a specific architecture (rows), trained on sparse grids with 40-line, 20-line, and 10-line spacing (columns). The figure visually demonstrates a key performance trade-off: while the standard U-Net and U-Net++ produce more complete surfaces, the Attention U-Net (Att-UNet) renders less complete horizons with larger gaps.}
    \label{fig:fig06}
\end{figure}

The standard U-Net and U-Net++ architectures operate as high-recall models, consistently producing more complete surfaces with coverage areas in the range $97$-$98\%$ (Table \ref{tab:results-f3}). However, this high recall can be associated with higher geometric errors (MAE of $4.38$ for U-Net and $4.21$ for Attn. U-Net at 40-line spacing). In contrast, Attention U-Net (Attn. U-Net) demonstrates a high precision, low recall strategy, consistently achieving the lowest MAE (ranging from $4.14$ to $4.21$) and MSE across all sparsity levels, indicating a high degree of geometric precision for the points it predicts. This precision comes at the cost of significantly lower surface coverage (recall), which drops to a low of $86.77\%$ at 40-line spacing. This behavior is qualitatively confirmed in Figure~\ref{fig:fig06}, which shows larger gaps in the surfaces predicted by the Attention U-Net.

Furthermore, the standard U-Net demonstrates superior robustness to data sparsity. As training data decreases from 10- to 40-line spacing, its validation IoU remains remarkably stable, dropping by only $0.019$ (from $0.946$ to $0.927$), while its surface coverage barely changes, declining by less than one percentage point (from $98.27\%$ to $97.40\%$). In contrast, the more complex models show significant performance degradation. The Attention U-Net’s validation IoU drops more steeply by $0.067 $ (from $0.918$ to $0.851$). This represents more than three times the degradation of the standard U-Net, while its coverage falls sharply by 6.5 percentage points ($93.26\%$ to $86.77\%$). The $CFA^{S}$ U-Net exhibits a similar drop in IoU, falling from $0.913$ to $0.857$. This suggests the simpler convolutional architecture of the standard U-Net possesses stronger generalization capabilities when data is limited.

\begin{figure}[t!]
    \centering
    \includegraphics[width=\linewidth]{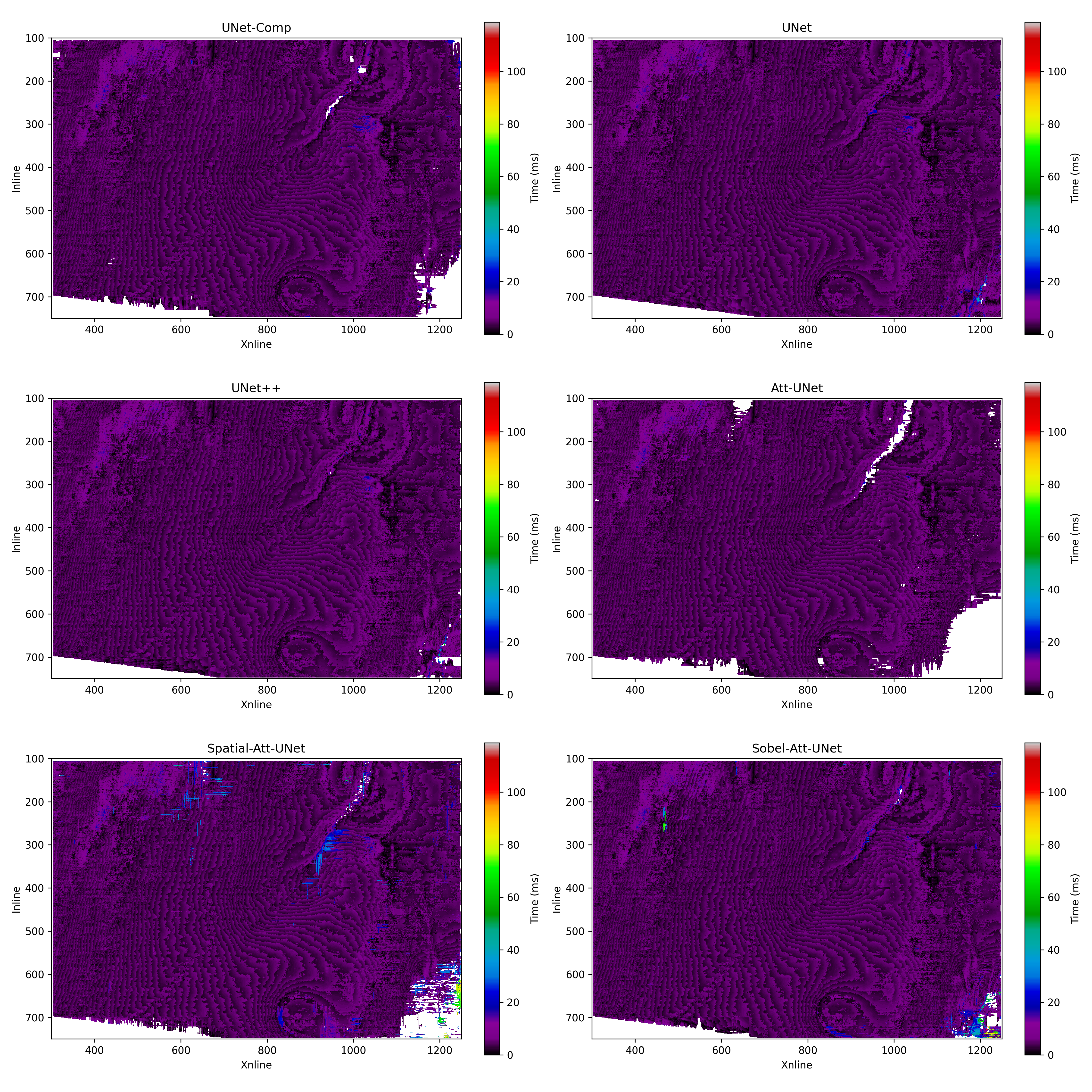}
    \caption{\textbf{Spatial distribution of prediction error (TWT difference) for the 10-line spacing scenario.} Each map shows the pixel-wise difference between the predicted and true two-way travel time. Cooler colors indicate minimal error, while warmer colors highlight regions with larger prediction discrepancies. Such visualizations help diagnose model performance, revealing that errors for most architectures are often concentrated along complex fault zones where horizon continuity is disrupted.}
    \label{fig:diff-f3}
\end{figure}

The proposed Context-Fusion Attention model was designed to address the characteristic high-precision, low-recall behavior of the standard Attention U-Net. The \textbf{$CFA^{S}$ U-Net} incorporates a learned $3 \times 3$ spatial head to encourage interpolation, while the full \textbf{$CFA$ U-Net} adds a Sobel filter head to provide explicit edge guidance. As shown in Table~\ref{tab:results-f3}, both models successfully improved surface coverage (recall), especially in sparse data conditions. At 40-line spacing, where the standard Attention U-Net coverage drops to $86.77\%$, the \textbf{$CFA^{S}$ U-Net} increases to $93.16\%$, a relative improvement of $7\%$. The \textbf{$CFA$ U-Net} further improves this metric to $97.61\%$, an increase of almost 11 percentage points over the Attention U-Net, effectively matching the high-recall performance of the best-in-class standard U-Net ($97.40\%$).

However, compared to the best-in-class MAE of the standard Attention U-Net of $4.14$ (with 10-line spacing), the \textbf{$CFA^{S}$ U-Net} registered a significantly higher MAE of $4.77$ (an increase of more than \textbf{$15\%$}). The addition of the Sobel head to \textbf{$CFA$ U-Net} mitigated this effect, reducing the MAE to $4.44$. Therefore, while the spatial head provides a strong inductive bias for surface completeness, the Sobel head constrains these interpolations geometrically. The \textbf{$CFA$ U-Net} represents the most effective trade-off, achieving near-optimal surface coverage while better preserving geometric precision than the spatial-only approach. This approach highlights the crucial need to evaluate both pixel-level and geometric metrics to understand model performance fully.

\begin{figure}[t!]
    \centering
    \includegraphics[width=1\linewidth]{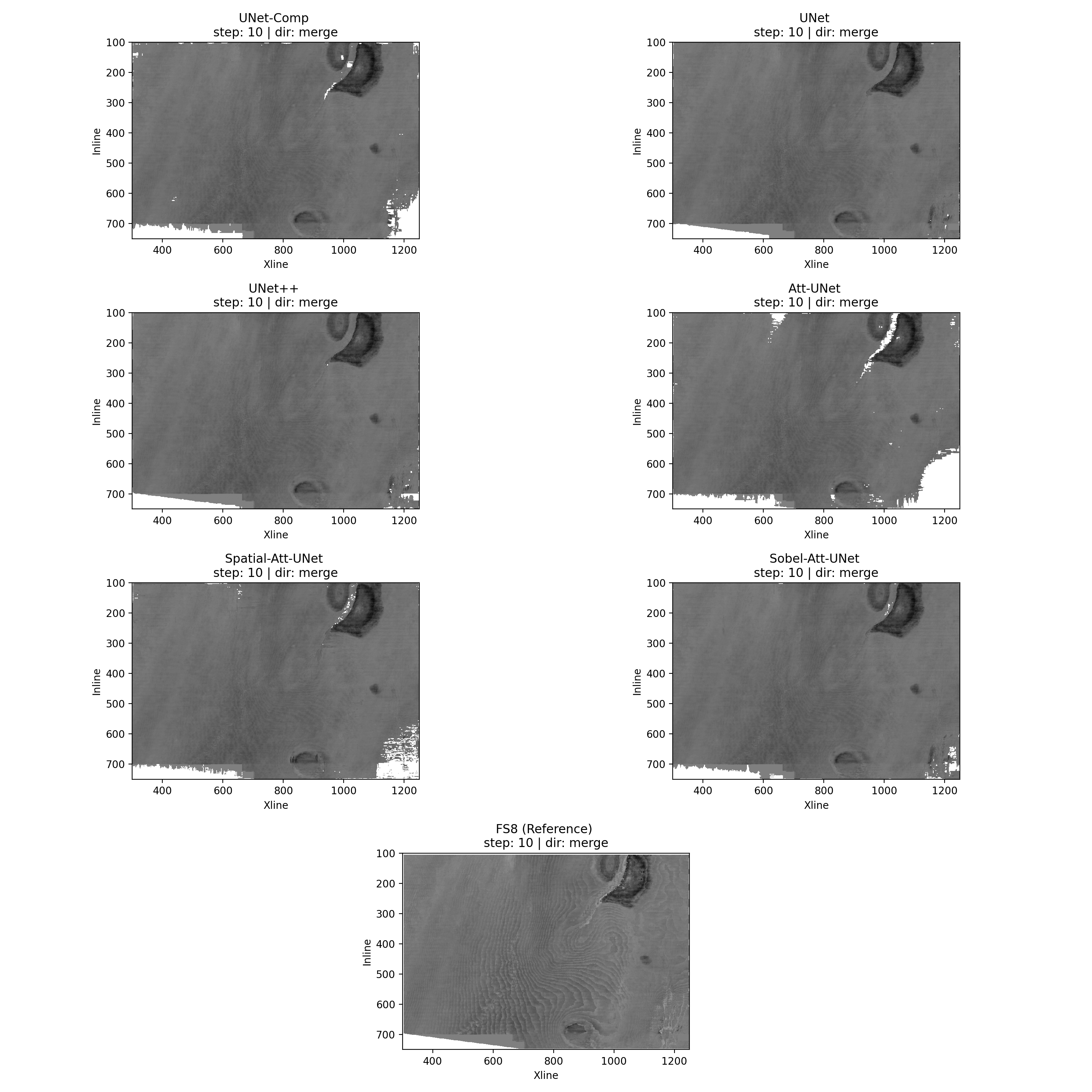}
    \caption{\textbf{Map view of predicted horizons on the F3 data set} with merged results for different architectures trained with a 10-line spacing for both inline and crossline directions. The bottom map shows the ground truth (FS8) horizon for reference.}
    \label{fig:results-mapview}
\end{figure}

Figure~\ref{fig:diff-f3} provides a qualitative complement to the metrics in Table~\ref{tab:results-f3} by visualizing the spatial distribution of prediction errors (TWT difference) for the 10-line spacing scenario. The error maps reveal the different strategies that the models employ when encountering geological uncertainty. The standard U-Net, with its broad convolutional feature extractors, interpolates across ambiguous regions such as fault zones where reflector continuity is low. This strategy prioritizes surface completeness (high recall), resulting in moderate, localized prediction errors, visualized as the cooler, blue regions representing differences of \textbf{10-20 ms}. Figure~\ref{fig:results-mapview} presents a map view of the merged horizons for each architecture trained with 10-line spacing in both directions. The observed gaps align with the model coverage reported in Table~\ref{tab:results-f3}.

In contrast, the Attention U-Net focuses only on high-confidence features. When its attention gate encounters the low-confidence signature of a fault zone, it correctly identifies this region as highly uncertain and actively suppresses the prediction entirely, creating a gap in the output surface. In the error map, these gaps manifest themselves as significant apparent differences (\textbf{$\sim$100 ms}), not because the model made an inaccurate prediction, but because it correctly identified its own uncertainty and abstained altogether. Therefore, in this case, the Attention U-Net achieves its superior MAE by precisely predicting only what it is sure about, forgoing risky interpolations in complex zones.

The introduction of the spatial-only head in the \textbf{$CFA^{S}$ U-Net} successfully improves the baseline Attention U-Net’s coverage from $93.26\%$ to $96.83\%$ (at 10-line spacing), but this comes with a sharp penalty to precision, as its MAE increases from $4.14$ to $4.77$. The full \textbf{$CFA$ U-Net}, which adds a Sobel filter head, demonstrates a superior solution. It enhances the surface coverage even further to \textbf{$98.19\%$}, nearly matching the high-recall standard U-Net ($98.27\%$), while the Sobel head constrains the geometry, mitigating the error and yielding a much lower MAE of \textbf{$4.44$}. Therefore, the \textbf{$CFA$ U-Net}  resolves the fundamental trade-off. It achieves the high surface coverage of a standard U-Net while maintaining geometric precision far superior to that of the spatial-only attention model, establishing it as the most balanced and effective architecture for horizon reconstruction in this data set.

\subsection{Mexilhão Field}

To assess architectural robustness, experiments were performed on the Mexilhão Field seismic volume, a data set characterized by significant faulting and lower reflector continuity. The evaluation again relies on comparing both pixel-level accuracy (IoU), which measures classification correctness, and post-processed geometric utility, which measures the real-world precision (MAE) and completeness (surface coverage) of the final horizon.

The increased geological complexity had a clear and quantifiable impact on performance compared to the F3 block. The standard U-Net's validation IoU at 10-line spacing, for example, dropped by \textbf{7.5\%} from $0.946$ on F3 to $0.875$ on Mexilhão (Table~\ref{tab:metrics-mexilhao}). This data set also inverted the relationship between geometric errors, as detailed in Table~\ref{tab:results-mexilhao}. While the Mean Absolute Error (MAE) for the U-Net was \textbf{40.2\%} lower (decreasing from 4.23 to 2.53), the Mean Squared Error (MSE) was \textbf{85.2\%} higher (increasing from 23.26 to 43.08). This divergence indicates that while predictions were often closer to the true horizon on average (lower MAE), they were also prone to larger, more significant outlier errors (higher MSE), which can be a direct consequence of the complex geology.

\begin{table}[t]
    \centering
    \begin{tabular}{|l|c|cc|cc|cc|}
        \toprule
        \hline
        \textbf{Architecture} & \textbf{Spacing} & \multicolumn{2}{c|}{\textbf{Mean Acc.}} & \multicolumn{2}{c|}{\textbf{Mean IoU}} & \multicolumn{2}{c|}{\textbf{Mean Dice Loss}} \\
        & & Train & Valid & Train & Valid & Train & Valid \\
        \hline
        \midrule
        \multirow{3}{*}{U-Net (Comp.)} & 10 & 0.9957 & 0.9942 & 0.8543 & 0.8063 & 0.4457 & 0.2848 \\
        & 20 & 0.9969 & 0.9946 & 0.8809 & 0.8070 & 0.3829 & 0.2954 \\
        & 40 & 0.9966 & 0.9929 & 0.8729 & 0.7643 & 0.4874 & 0.3675 \\
        \hline
        \multirow{3}{*}{U-Net} & 10 & 0.9993 & 0.9968 & 0.9685 & 0.8747 & 0.0332 & 0.1439 \\
        & 20 & 0.9995 & 0.9968 & 0.9769 & 0.8724 & 0.0240 & 0.1468 \\
        & 40 & 0.9987 & 0.9954 & 0.9432 & 0.8240 & 0.0627 & 0.2165 \\
        \hline
        \multirow{3}{*}{U-Net++} & 10 & 0.9997 & 0.9963 & 0.9877 & 0.8573 & 0.0158 & 0.1690 \\
        & 20 & 0.9997 & 0.9962 & 0.9866 & 0.8515 & 0.0225 & 0.1791 \\
        & 40 & 0.9996 & 0.9948 & 0.9825 & 0.8086 & 0.1070 & 0.2813 \\
        \hline
        \multirow{3}{*}{Attn. U-Net} & 10 & 0.9998 & 0.9963 & 0.9899 & 0.8500 & 0.0122 & 0.1833 \\
        & 20 & 0.9997 & 0.9959 & 0.9890 & 0.8322 & 0.0385 & 0.2365 \\
        & 40 & 0.9998 & 0.9944 & 0.9893 & 0.7770 & 0.0642 & 0.3500 \\
        \hline
        \multirow{3}{*}{\textbf{$CFA^{S}$ U-Net}} & 10 & 0.9998 & 0.9963 & 0.9911 & 0.8478 & 0.0154 & 0.1938 \\
        & 20 & 0.9997 & 0.9955 & 0.9875 & 0.8162 & 0.0459 & 0.2731 \\
        & 40 & 0.9997 & 0.9944 & 0.9882 & 0.7755 & 0.0780 & 0.3592 \\
        \hline
        \multirow{3}{*}{\textbf{$CFA$ U-Net}} & 10 & 0.9996 & 0.9971 & 0.9827 & 0.8812 & 0.0213 & 0.1404 \\
        & 20 & 0.9996 & 0.9968 & 0.9841 & 0.8695 & 0.0195 & 0.1554 \\
        & 40 & 0.9995 & 0.9959 & 0.9789 & 0.8386 & 0.0449 & 0.2201 \\
        \hline
        \bottomrule
    \end{tabular}
    \caption{\textbf{Segmentation metrics on the Mexilhão data set.} "Valid" refers to the validation set. Values represent the mean of models trained on inline and crossline data. U-Net (Comp.) and Attn. U-Net represents Compressed U-Net and Attention U-Net, respectively.}
    \label{tab:metrics-mexilhao}
\end{table}

The challenging geology of the Mexilhão Field reveals a sharp trade-off between geometric precision and surface completeness. The results, summarized in Table~\ref{tab:results-mexilhao}, show that no single architecture dominates on both metrics. Instead, a clear hierarchy emerges where the proposed \textbf{$CFA$ U-Net} excels at precision, while the \textbf{U-Net++} architecture is the unambiguous leader in recall.

\begin{table}[h!]
    \centering
    \begin{tabular}{lccccc}
        \toprule
        \textbf{Architecture} & \textbf{Spacing} & \textbf{MAE} & \textbf{MSE} & \textbf{Area (\%)} \\
        \midrule
        \multirow{3}{*}{U-Net (Comp.)} & 10 & 2.89 & 60.79 & 94.43 \\
        & 20 & 2.85 & 61.32 & 91.03 \\
        & 40 & 3.93 & 126.51 & 91.17 \\
        \hline
        \multirow{3}{*}{U-Net} & 10 & 2.53 & 43.08 & 95.86 \\
        & 20 & 2.64 & 48.52 & 95.41 \\
        & 40 & 3.11 & 72.14 & 94.27 \\
        \hline
        \multirow{3}{*}{U-Net++} & 10 & 2.67 & 52.87 & 96.25 \\
        & 20 & 2.93 & 66.53 & 95.82 \\
        & 40 & 4.17 & 146.23 & 94.79 \\
        \hline
        \multirow{3}{*}{Attn. U-Net} & 10 & 2.73 & 59.10 & 90.35 \\
        & 20 & 3.21 & 93.55 & 87.30 \\
        & 40 & 3.56 & 119.03 & 78.70 \\
        \hline
        \multirow{3}{*}{\textbf{$CFA^{S}$ U-Net}} & 10 & 2.67 & 55.37 & 89.96 \\
        & 20 & 3.48 & 117.55 & 85.15 \\
        & 40 & 3.32 & 98.92 & 79.45 \\
        \hline
        \multirow{3}{*}{\textbf{$CFA$ U-Net}} & 10 & 2.49 & 45.30 & 92.97 \\
        & 20 & 2.94 & 72.18 & 92.12 \\
        & 40 & 3.42 & 106.97 & 89.34 \\
        \bottomrule
    \end{tabular}
    \caption{Geometric errors (MAE, MSE) and surface coverage area of the merged and filtered results for each model on the Mexilhão data set, including the proposed $CFA^{S}$ U-Net with spatial head and $CFA$ U-Net including semantic, spatial, and sobel heads.}
    \label{tab:results-mexilhao}
\end{table}

On this challenging data set, the proposed \textbf{$CFA$ U-Net emerges as the top-performing architecture in terms of overall precision} when the data is reasonably dense. At the 10-line spacing, it achieved both the highest validation IoU ($0.881$) and the lowest MAE ($2.49$), demonstrating superior pixel-level and geometric accuracy (Tables \ref{tab:metrics-mexilhao} and \ref{tab:results-mexilhao}). This MAE is \textbf{1.6\%} lower than the standard U-Net's ($2.53$) and \textbf{6.7\%} lower than the U-Net++'s ($2.67$). This result is significant, as it shows that the multi-head attention mechanism, which provides spatial and edge-aware inductive biases, is highly effective at navigating the complex faulting that challenges other models.

\begin{figure}[t!]
    \centering
    \includegraphics[width=1\linewidth]{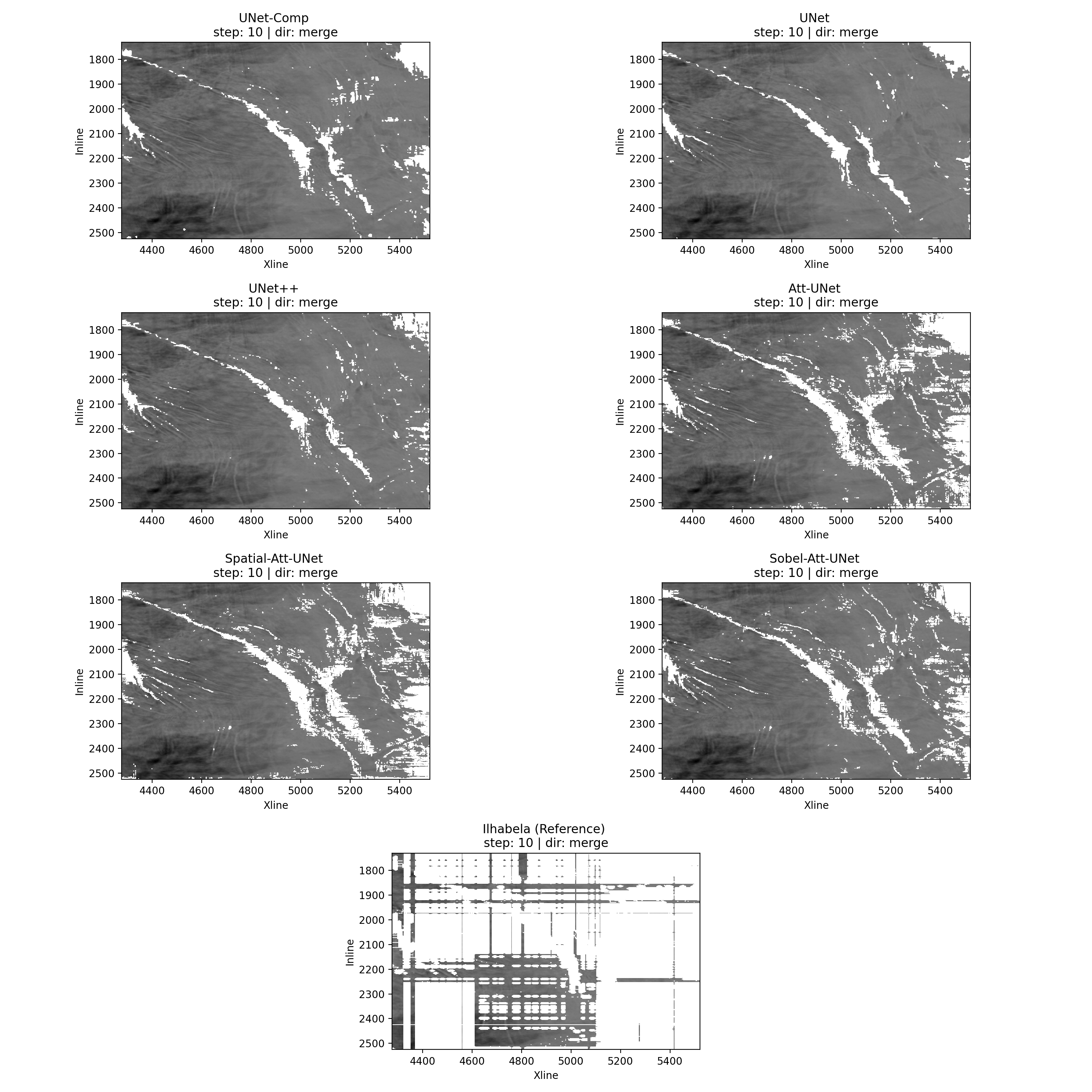}
    \caption{\textbf{Map view of predicted horizons on the Mexilhão data set.} The panel shows the merged results for each architecture trained with a 10-line spacing for both directions (the inline and cross-line), as indicated by its title. Models (a) Compressed U-Net, (b) U-Net, (c) U-Net++, (d) Attention U-Net, (e) Spatial CFA U-Net, (f) Context-Fusion Attention (Sobel, Spatial and Context) U-Net, and (g) the ground truth interpretation, highlighting the irregular grid used for training samples.}
    \label{fig:results-mexilhao}
\end{figure}

While the $CFA$ U-Net was the most precise, the \textbf{U-Net++ architecture proved to be the best approach for recall}, consistently delivering the most complete surfaces, as quantified in Table~\ref{tab:results-mexilhao} and visualized in Figure~\ref{fig:results-mexilhao}. At 10-line spacing, it achieved a surface coverage of \textbf{96.25\%}, the highest of any model, which is a notable improvement over the standard U-Net ($95.86\%$) and significantly higher than the high-precision $CFA$ U-Net ($92.97\%$). However, this superior recall came at the cost of precision, with an MAE of $2.67$ that was higher than both the standard U-Net and the $CFA$ U-Net. In contrast, the standard \textbf{Attention U-Net produced} one of the least complete surfaces (coverage of $90.35\%$) without achieving a competitive level of geometric precision (MAE of $2.73$), highlighting its limitations in regions of widespread geological ambiguity.

\begin{figure}[h!]
    \centering
    \includegraphics[width=1\linewidth]{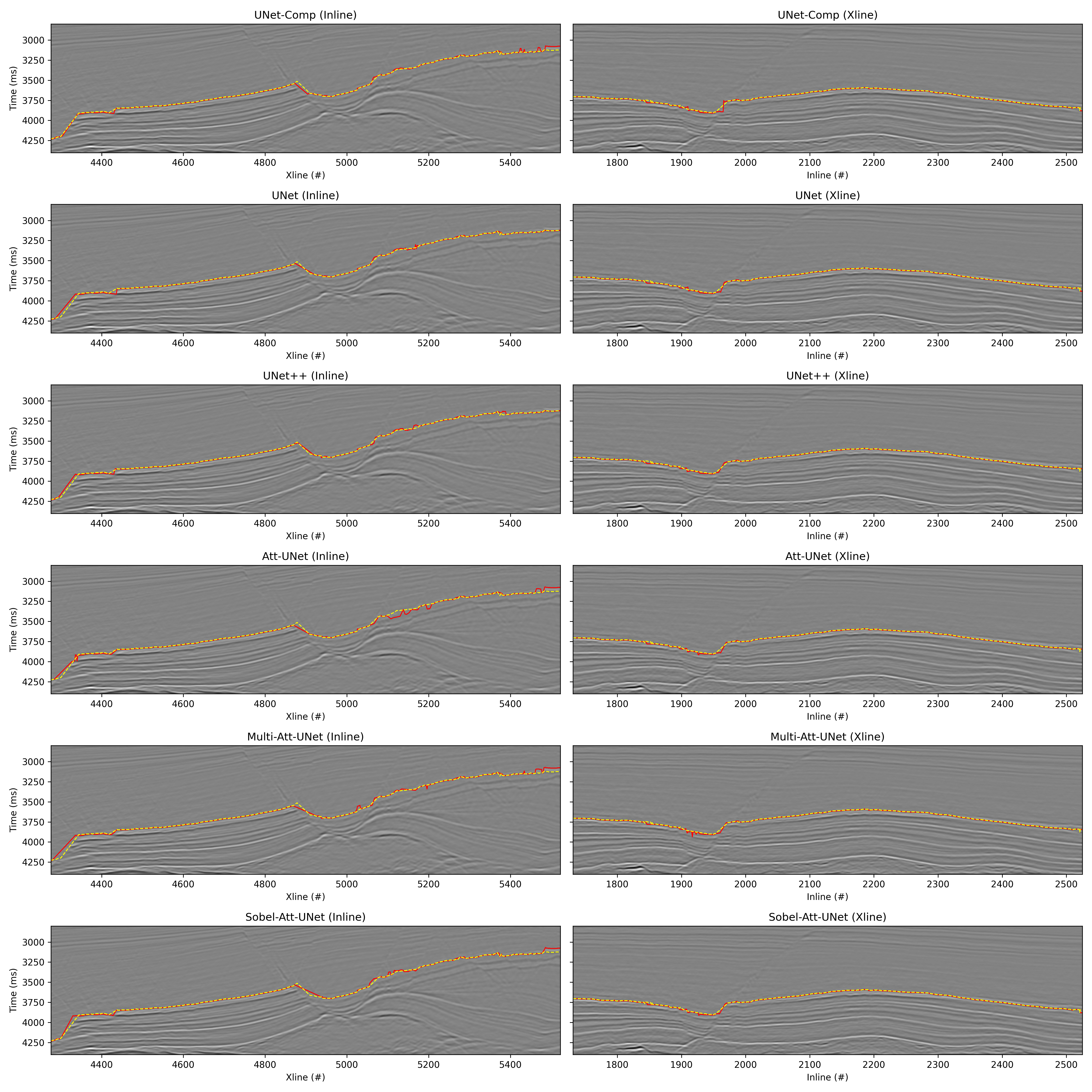}
    \caption{\textbf{2D profile comparison of predicted horizons}. This figure shows a cross-section of the results on a specific inline (left) and crossline  (right). The predicted horizon (red line) from each architecture is compared against the true interpreted horizon (dotted yellow line).}
    \label{fig:results-line-mexilhao}
\end{figure}

A more detailed, cross-sectional view of these predictions is provided in Figure~\ref{fig:results-line-mexilhao}. This 2D profile view allows for a direct comparison of the predicted horizon (red line) from each architecture against the ground truth (dotted yellow line). These profiles visually corroborate the quantitative metrics, illustrating how the predictions from models like U-Net++, the standard U-Net, Attention U-Net, and both Context-Fusion Attention U-Net closely follow the true horizon across large continuous segments, while also revealing the small-scale geometric deviations that contribute to their respective MAE scores.

The performance hierarchy changes significantly under the most challenging condition of extreme data sparsity (40-line spacing), as detailed in Table~\ref{tab:results-mexilhao}. As the training set is reduced to every 40th line, the architectural complexity that benefited $CFA$ U-Net becomes a liability. Its geometric precision declines, with its MAE increasing to $3.42$. In this low-data regime, the simpler \textbf{standard U-Net demonstrates superior robustness, reclaiming the top spot for geometric precision with the lowest MAE of 3.11}. This result is nearly \textbf{10\%} better than the $CFA$ U-Net's MAE, suggesting that the standard U-Net's less complex convolutional structure provides a more substantial, more generalizable inductive bias when training data is severely limited. Even in this sparse scenario, U-Net++ maintains its status as the recall leader with a coverage of $94.79\%$.

In conclusion, while the sophisticated attention mechanism of the \textbf{$CFA$ U-Net} makes it the best model for navigating complex geology with adequate data, the architectural simplicity of the \textbf{standard U-Net} makes it more robust and reliable under conditions of extreme data scarcity. The \textbf{U-Net++} remains the most practical choice in any scenario where maximizing surface continuity was the primary objective. This nuanced, task-dependent hierarchy highlights the critical importance of selecting an architecture based not only on the geological setting but also on the density of the available training data.

\subsection{Discussion}

The U-Net-based architectures have been systematically evaluated, confirming that the optimal choice for seismic interpretation is highly dependent on both geological complexity and data sparsity. The evidence is found in the performance drop between the simple F3 block and the complex Mexilhão field, where the peak validation IoU for the best-performing model fell from $0.946$ (Table~\ref{tab:metrics-f3}) to $0.881$ (Table~\ref{tab:metrics-mexilhao}). The data scarcity further compounded this challenge, as seen on the complex Mexilhão data where the standard U-Net's geometric error (MAE) increased by nearly $23\%$ when the training data was reduced (Table~\ref{tab:results-mexilhao}).

The analysis of the established architectures revealed distinct, specialized roles. The standard U-Net proved to be a robust all-rounder, while U-Net++ consistently delivered the highest surface coverage (e.g., $96.3\%$ on Mexilhão, Table~\ref{tab:results-mexilhao}), making it the best choice for large-scale structural mapping. The Attention U-Net confirmed its status as a high-precision specialist on simple data, achieving the lowest geometric error (MAE of $4.14$~ms) on the F3 data set (Table~\ref{tab:results-f3}). However, its cautious prediction strategy, which relies on high-confidence features, proved to be a critical weakness in complex geology, a trade-off visually apparent in the surface gaps shown in Figure~\ref{fig:fig06}. On the Mexilhão data set, its surface coverage fell to a low of $78.7\%$ (Table~\ref{tab:results-mexilhao}) as the attention gates learned to suppress predictions in faulted zones.

The design of the proposed Context-Fusion Attention (CFA) U-Net was validated through an ablation study. An intermediate model with only a spatial head ($CFA^{S}$ U-Net) successfully addressed the low-recall problem of the standard Attention U-Net, increasing coverage from $93.3\%$ to $96.8\%$ at 10-line spacing. However, this came at the cost of a sharp increase $15\%$ in geometric error and the highest MAE scores, which reached a peak at $5.18$~ ms (Table~\ref{tab:results-f3}). This finding proved that adding a spatial prior alone is insufficient, thus validating the need for the Sobel head in the full CFA U-Net to create a more balanced architecture.

The success of the full CFA U-Net was first demonstrated by its class-leading robustness in the F3 data set. It completely resolved the recall issue of the baseline attention model, boosting its surface coverage at 40-line spacing by a massive 10.8 percentage points. This improvement was so significant that it achieved the highest surface coverage ($97.6\%$) of all architectures in this data-sparse scenario, outperforming even the high-recall of U-Net++ ($97.3\%$), as shown in Table~\ref{tab:results-f3} and visualized in Figure~\ref{fig:fig06}. Therefore, the context-aware design provides a superior inductive bias, allowing the model to generalize exceptionally well from limited data.

The performance of the CFA U-Net was fully assessed on the challenging Mexilhão data set. The CFA U-Net enhanced precision, providing both the highest IoU validation ($0.881$, Table~\ref{tab:metrics-mexilhao}) and the lowest MAE ($2.49$~ms, Table~\ref{tab:results-mexilhao}) of any model. This dual achievement, visually corroborated by the map views and 2D profiles in Figures~\ref{fig:results-mexilhao} and \ref{fig:results-line-mexilhao}, confirms that its hybrid attention mechanism is uniquely effective at navigating geologically complex regions. Therefore, the validation of the CFA U-Net and the hybrid workflow shows that context-aware attention mechanisms are critical components to creating more robust, accurate, and practical tools for automated seismic interpretation.

\subsection{Conclusion}

The results established that while standard models like U-Net and U-Net++ are robust baselines that excel at all-around performance and high-recall mapping, respectively, the standard Attention U-Net is a high-precision specialist whose practical utility is limited by poor surface completeness on complex geology. This precision-recall trade-off motivated the development of our proposed context-aware architecture.

The CFA U-Net successfully addressed these limitations, demonstrating class-leading robustness on the F3 data set. The model improved the critical recall issue of the baseline attention model, boosting surface coverage at 40-line spacing by a massive 10.8 percentage points. Furthermore, the improvement was so significant that it achieved the highest surface coverage ($97.6\%$) of all architectures in this data-sparse scenario, outperforming even the high-recall champion, U-Net++ ($97.3\%$). Therefore, the context-aware design provides a superior inductive bias for generalizing from limited data.

Furthermore, the CFA U-Net's capabilities were also assessed on the challenging Mexilhão data set. The results demonstrate high precision, yielding both the highest validation IoU ($0.881$) and the lowest MAE ($ 2.49$ ms) among all models. Therefore, the hybrid fusion attention mechanism is uniquely effective at navigating geologically complex regions to produce the most accurate and reliable interpretations as compared to the Attention U-Net.

Ultimately, this robust hybrid workflow successfully combines neural networks with DBSCAN and merged orthogonal (inline and crossline) predictions for comprehensive 3D horizon reconstruction. The development of context-aware attention mechanisms was validated through the proposed CFA U-Net, which is considered a promising path toward creating practical interpretation tools.

\newpage
\section*{Data and Materials Availability}

Data associated with this research are available and can be obtained by contacting the corresponding author.

\section*{Corresponding Author}

\noindent Correspondence and requests for materials should be addressed to 
\textit{Dr.Jose Luis Silva} at \texttt{jseluis.silva@gmail.com}.

\section*{Acknowledgements}

This work was supported by the National Council for Scientific and Technological 
Development (CNPq), Brazil, under grants No. 409718/2022-0 and No. 445344/2024-5.